\documentclass[11pt]{article}
\pdfoutput=1

\usepackage{acl}
\usepackage{times}
\usepackage{lineno}
\usepackage{hyperref}
\usepackage{tabularx}
\usepackage{float}

\usepackage{latexsym}
\usepackage[T1]{fontenc}

\usepackage[utf8]{inputenc}

\usepackage{microtype}

\usepackage{inconsolata}

\usepackage{graphicx}
\usepackage{svg}
\usepackage{multirow}
\usepackage{booktabs}

\usepackage{caption}
\usepackage{subcaption}
\usepackage{enumitem}
\usepackage{float}
\usepackage{amsmath}
\usepackage{amssymb}
\usepackage[dvipsnames]{xcolor}

\usepackage{amsmath,amsfonts,bm}

\def\eqref#1{equation~\ref{#1}}

\def\1{\bm{1}}

\DeclareMathAlphabet{\mathsfit}{\encodingdefault}{\sfdefault}{m}{sl}
\SetMathAlphabet{\mathsfit}{bold}{\encodingdefault}{\sfdefault}{bx}{n}

\newcommand{\softmax}{\mathrm{softmax}}

\newcommand{\methodname}{AtManRL}

\title{\methodname: Towards Faithful Reasoning \\ via Differentiable Attention Saliency} 

\author{Max Henning Höth \\
  Aleph Alpha Research\\ Lab1141 \\ \\\And
  Kristian Kersting \\
  TU Darmstadt \\ Hessian.AI \\ Lab1141 \\ \\\And
  Björn Deiseroth \\
  Aleph Alpha Research\\ Lab1141 \\ \\\And
  Letitia Parcalabescu \\
  Aleph Alpha Research \\ Lab1141 \\
  }

\begin{document}

\maketitle

\begin{abstract}
Large language models (LLMs) increasingly rely on chain-of-thought (CoT) reasoning to solve complex tasks, yet a reasoning trace may merely accompany the final answer rather than causally contribute to it or faithfully reflect how it was reached. We introduce differentiable attention manipulation to train salient reasoning with reinforcement learning -- method which we call \textsc{AtManRL}. By optimizing an additive attention mask that identifies the CoT tokens needed to recover the model's answer, we derive a saliency reward and combine it with an outcome reward in GRPO with decoupled advantages. Because both saliency and faithfulness depend strongly on trace length, we evaluate against a sweep of length-penalty baselines that traces the trade-off a pure length penalty can reach, rather than against a single baseline. Across three models (Qwen2.5-3B, Granite-3.3-2B, and Phi-4-mini) and four benchmarks (GSM8K, MMLU, ProofWriter, and QuestBench), \textsc{AtManRL} sits above this trade-off curve by 3.9-4.0pp in saliency, measured as answer changes under mistake injection, and by 5.7-7.7pp in faithfulness, measured as hint verbalization under hint injection, while matching the accuracy of baselines of similar length. These results suggest that training explicitly for saliency is a promising route towards chain-of-thought reasoning that is more faithful to the model's actual decision process.

\end{abstract}

\section{Introduction}
\label{sec:intro}
The reasoning abilities of large language models (LLMs) have been steadily improved by
Chain-of-thought (CoT) prompting \citep{wei_chain--thought_2022}, supervised learning, and reinforcement learning (RL) approaches \citep{yang2025qwen3technicalreport, openai2024openaio1card, Guo_2025}. 
By generating intermediate reasoning steps before the final answer, models often reach higher accuracy on complex tasks. 

The presence of a reasoning trace, however, does not guarantee that the model actually uses it to arrive at its answer. Consequently, 
a central question is: 
\emph{Does the generated CoT causally influence the model’s final prediction and has explanatory power, or does it merely accompany it as a stylistic artifact?}
This question relates to the notion of \emph{faithfulness}, which asks whether an explanation reflects the model’s true decision-making process \citep{jacovi2020faithfullyinterpretablenlpsystems}. 
An unfaithful reasoning trace may appear \textit{plausible} and logically coherent while the model reaches the correct answer through shortcuts that bypass the stated reasoning \citep{agarwal2024faithfulness}. 
Prior work shows that LLMs can produce plausible-sounding CoT explanations that do not align with the mechanisms that drive their predictions \citep{turpin2023language, lanham_measuring_2023, barez2025chain}.
To investigate this gap, we distinguish between \emph{saliency} and \emph{faithfulness}: 

\textbf{Saliency.} We define saliency as the measurable causal contribution of individual reasoning tokens to the final answer logits. 
Ensuring \textit{saliency of the reasoning trace} prevents CoT from degenerating into lengthy yet weakly relevant narratives. Without such constraints, reasoning traces risk functioning as post-hoc rationalizations rather than interpretable evidence of the model’s computation.

\textbf{Faithfulness.} A faithful reasoning trace must accurately reflect the true reasoning process of the model \cite{jacovi2020faithfullyinterpretablenlpsystems}. When unfaithful, the model could produce a convincing sounding reasoning trace which does not reflect the latent computation at all.
Saliency constitutes a necessary but not sufficient condition for faithfulness.

Guided by this distinction, we propose \textsc{\methodname}, a method that explicitly trains models to produce salient reasoning traces using reinforcement learning. Our approach builds on \textsc{AtMan} \citep{deiseroth_atman_2023}, an attention manipulation technique that allows targeted modification of attention weights through a predefined mask. Whereas prior work uses \textsc{AtMan} for post-hoc interpretability, we instead treat the attention manipulation mask as a learnable, differentiable object. This allows us to: (i) \emph{efficiently} identify which tokens in the reasoning trace are truly influential for the final answer, (ii) derive a saliency-based reward signal from these contributions, and (iii) incorporate this signal into reinforcement learning to encourage the generation of salient reasoning steps while discouraging extraneous or weakly relevant explanatory content.

Overall, our contributions are: (1) We introduce a \emph{saliency reward} derived from optimizing a differentiable attention mask that identifies salient tokens in the CoT, namely the \textsc{\methodname} reward. (2) We combine this reward with outcome-based rewards in the GRPO framework to \emph{jointly train for correctness and saliency}. (3) We evaluate our method on GSM8K, MMLU, Questbench\_GSME, and Proofwriter using Phi-4-mini-instruct, Granite-3.3-2B-Instruct, and Qwen2.5-3B-Instruct and show that we can \textit{reduce length, preserve accuracy, increase faithfulness, and increase saliency}. (4) Because faithfulness and trace length are empirically entangled, we evaluate under length stratification and against a length-penalty baseline sweep rather than a single pooled comparison.

\section{Related Work}

\textbf{CoT / Reasoning Traces.}
CoT prompting \citep{wei_chain--thought_2022} and RL methods like GRPO \citep{shao2024deepseekmath} encourage LLMs to generate reasoning traces. RL optimizes based on outcome rewards. 
However, they focus on answer correctness and do not explicitly enforce causal dependency between reasoning trace and final prediction. 
In contrast, we explicitly reward causal dependency between CoT tokens and the answer.

\textbf{Reasoning Trace Faithfulness.}
The faithfulness of model explanations has been studied extensively in interpretability research. \citet{lanham_measuring_2023} introduced a family of intervention tests, such as early answering, mistake injection, paraphrasing, and filler substitution, to probe whether a manipulated CoT causally influences the final answer. Expanding on how models are swayed by unmentioned biasing features \citep{turpin2023language}, hint-based tests measure whether a model verbalizes an injected hint that changed its answer \citep{chua2025deepseekr1reasoningmodels, chen2025reasoningmodelsdontsay}. We follow the formulation of \citet{chen2025reasoningmodelsdontsay}. With such tests, reasoning models were found to verbalize hints more often than non-reasoning models \citep{chua2025deepseekr1reasoningmodels}, yet unfaithful reasoning also occurs on realistic prompts without any injected bias \citep{arcuschin2026chainofthought}. Other work measures faithfulness through causal mediation analysis \citep{paul2024makingreasoningmattermeasuring} or by unlearning individual reasoning steps from the model parameters \citep{tutek2025measuringchainthoughtfaithfulness}. \citet{parcalabescu-frank-2024-measuring} question whether edit-based tests capture faithfulness to the model's computation or only self-consistency, \citet{bentham2024chainofthoughtunfaithfulnessdisguisedaccuracy} question the validity of the CoT-dependence metric used in the scaling experiments of \citet{lanham_measuring_2023}, showing that it correlates strongly with accuracy once corrected for answer-choice bias, and \citet{zaman2026chainofthoughtreallyexplainabilitychainofthought} argue that hint verbalization conflates unfaithfulness with incomplete explanations, showing that it increases with the available token budget. Conversely, \citet{chen2025reasoningmodelsdontsay} find that unfaithful CoTs of reasoning models tend to be longer than faithful ones, so length and faithfulness are entangled rather than independent.

Unlike \citet{chen2025reasoningmodelsdontsay}, we evaluate small instruction-tuned models with traces orders of magnitude shorter, comparing models optimized for different lengths rather than sampling from a single model. Broader critiques argue that CoT should not be conflated with explainability at all \citep{barez2025chain}. In light of these critiques, we use mistake injection \citep{lanham_measuring_2023} to measure saliency rather than faithfulness, and hint injection \citep{chen2025reasoningmodelsdontsay} to measure faithfulness, comparing methods at matched trace length and reporting accuracy alongside.

A smaller line of work trains models towards more faithful reasoning: FRODO \citep{paul2024makingreasoningmattermeasuring} uses counterfactual and causal preference objectives so that the answer follows the stated reasoning, and verbalization fine-tuning \citep{turpin2025teachingmodelsverbalizereward} trains models to acknowledge prompt hints that influence them. In contrast, we reward the causal dependence of the answer on the trace directly through an attention-based saliency signal, and optimize for neither of our evaluation metrics.

\textbf{Critical Reasoning Tokens.}
Prior work showed that individual CoT tokens (called critical tokens) can decide whether a reasoning trace reaches the correct answer. \citet{vassoyan_ignore_2025} encouraged exploration on such tokens to improve RL fine-tuning efficiency. \citet{yan2025dont} intervene on attention weights to mitigate over-reliance on misleading tokens in few-shot examples. At the sentence level, \citet{bogdan2025thought} identify thought anchors, reasoning steps that disproportionately shape the rest of the trace, using counterfactual resampling, attention aggregation, and attention suppression between sentences. Whereas these methods use step importance to analyse traces or to steer credit assignment, exploration, or inference-time attention, we learn token-level saliency through differentiable attention manipulation and use it as a reward, so that RL shapes the trace itself.

\textbf{Attention-based Interpretability.}
Whether attention weights are useful for explainability is debated \citep{jain2019attentionexplanation, wiegreffe-pinter-2019-attention, serrano-smith-2019-attention}, which has motivated attribution by intervention rather than by reading off attention weights. Perturbation-based attribution removes or masks inputs and measures the effect on the prediction \citep{lundberg2017}. ERASER \citep{deyoung-etal-2020-eraser} formalizes this as comprehensiveness and sufficiency. Instead of enumerating perturbations, a mask can also be learned: \citet{Fong_2017} optimize the smallest image region whose deletion reduces a classifier's score, and DiffMask \citep{de-cao-etal-2020-decisions} learns a differentiable mask that removes as much of the input or hidden states as possible while preserving the prediction. \textsc{AtMan} \citep{deiseroth_atman_2023} perturbs the model through its attention instead, suppressing individual tokens in the pre-softmax scores without altering input or positional embeddings. We treat the \textsc{AtMan} suppression as a learnable, differentiable mask over CoT tokens and optimize it with gradient descent to restore the model's own answer likelihood. \textsc{AtManRL} applies it \textit{during training} by transforming the learned mask into an RL reward.

\textbf{Length Penalties.}
\citet{pfau2024letsthinkdotdot} shows that longer CoTs are not necessarily a more informative and that models trained with content-free filler tokens can use them as additional computation on certain tasks. RL-based methods for efficient reasoning commonly add a length penalty as auxiliary reward. Kimi~k1.5 \citep{kimiteam2025kimik15scalingreinforcement} applies it to all rollouts: correct ones receive a bonus or penalty relative to the group's length range, while incorrect ones receive only the penalty, so they too are pushed towards shorter lengths. \citet{zhang2025grpoleaddifficultyawarereinforcementlearning, arora2025traininglanguagemodelsreason, deng2026rearl, liu2025learnreasonefficientlyadaptive} instead apply the penalty only to correct rollouts, which \citet[App.~D]{liu2025learnreasonefficientlyadaptive} find preserves accuracy better. We follow this design for our length penalty baselines.

\section{Differentiable Attention Manipulation for Faithful
Reasoning}
\label{method}
We introduce our method \textsc{\methodname} to train models to produce salient reasoning traces by framing \textsc{AtMan} as a differentiable attention mask. 
Specifically, we (1) recap \textsc{AtMan}, (2) describe how we learn the mask, (3) derive a saliency measure from the optimized mask, and finally, (4) integrate saliency as a RL reward during training.
\subsection{Background: \textsc{AtMan} Attention Manipulation}
First, we review the additive \textsc{AtMan}-attention manipulation introduced in \cite{deiseroth_atman_2023}.
In a standard transformer, attention outputs are computed as: $O = \softmax(H)\cdot V$, where $\cdot$ denotes matrix multiplication and where the pre-softmax attention scores are given by
$H = Q K^T / \sqrt{d}$. Here, $Q, K, V \in \mathbb{R}^{h \times s \times d}$ denote the query, key, and value tensors with $h$ attention heads, sequence length $s$, and head dimension $d$.
Atman manipulates the pre-softmax scores $H$ with an additive mask  $H^\textsc{AtMan} \in \mathbb{R}^{s \times s}$:
\begin{equation}
    H = Q \cdot K^T / \sqrt{d} + H^\textsc{AtMan}
    \label{eqn:attention}
\end{equation}
Applying the mask $H^\textsc{AtMan}$ before the softmax ensures that the resulting attention scores still add to one after the softmax. Additionally, unlike other perturbation methods in XAI (e.g. SHAP \cite{lundberg2017}), this does not introduce a shift in the input-distribution or positional embeddings, but carefully manipulates the attention of the model of every single token. Positive mask values increase attention to selected tokens, whereas negative values suppress their influence. 
For autoregressive models, we additionally apply a causal bias $T \in \{0, -\inf\}^{s\times s}$, with $T_{ij} = 0$ for $j \le i$ and $T_{ij} = -\inf$ for $j > i$ and compute $H_M = H + T$. \citet{deiseroth_atman_2023} used $H^\textsc{AtMan}$ to suppress the attention to individual tokens by assigning a fixed negative value -- treated as a hyperparameter -- to the corresponding columns of  $H^\textsc{AtMan}$ to analyse each individual token's impact on the output logits of the LLM. However, we will treat $H^\textsc{AtMan}$ as a differentiable mask.
\subsection{Training an $H^{\textsc{AtMan}}$ Mask for Measuring Saliency}
\label{sec:learn_mask}
\begin{figure*}[t]
    \centering
    \includegraphics[width=\linewidth,trim={1cm 0 0.5cm 0},clip]{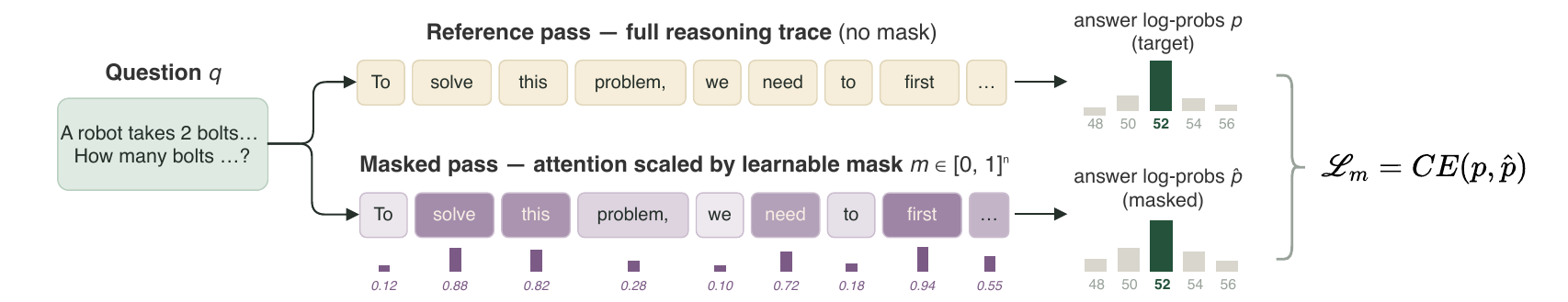}
    \caption{To identify non-salient tokens, we optimize the mask per rollout with the goal to restore the predicted answer log probabilities.}
    \label{fig:mask_loss}
\end{figure*}

We restrict  $H^{\textsc{AtMan}}$ to tokens within the reasoning trace (CoT) and leave the attention of prompt tokens or final answer tokens unmodified.
The prompt is outside the model’s control and therefore not a target for reward shaping. 
The final answer, however, should depend causally on the reasoning trace.
If the reasoning trace is salient, perturbing its attention should affect the probability of the model's answer. Since the mask enters the pre-softmax attention scores additively, it is fully differentiable.
We learn one value per CoT token (key), $m \in [0, 1]^T$, and multiply it by a negative constant $c$. This $m_v = 1$ suppresses token $v$, $m_v = 0$ leaves it unperturbed, and the  $[0, 1]$ range allows only suppression, never amplification. We tile the one mask key value $m_{v}$ across query rows:
\begin{equation}
H^{\textsc{AtMan}}_{T,v} = c\, m_v, \qquad v \in \{1,\dots,T\}.
\end{equation}
and leave entries outside the CoT span unmodified. The reward, therefore, only needs the query row of the last CoT token.
We initialize $m = \mathbf{1}$, which uniformly suppresses attention of reasoning tokens.
We then optimise $m$, the only trainable object, to restore the probability of the model's own answer tokens $y_{1:N}$ under teacher forcing (Fig.~\ref{fig:mask_loss}).
\begin{flalign}
\label{eqn:atman_loss}
& \mathcal{L}_{\text{mask}} = -\frac{1}{N}\sum\nolimits_{n=1}^{N} \nonumber \\
& \log P\big(y_n \mid c_{1:T}, y_{1:n-1}, H^{\textsc{AtMan}}\big),
\end{flalign}
where $c_{1:T}$ denotes the CoT tokens. Setting $m =\mathbf{0}$ recovers the unperturbed loss $\mathcal{L}_{\text{unmasked}}$. Optimization stops per rollout once a fraction of the loss has been recovered, where $\rho$ is the convergence ratio:
\begin{equation}
\frac{\mathcal{L}_{t^*} - \mathcal{L}_{\text{unmasked}}}{\mathcal{L}_{t=0} - \mathcal{L}_{\text{unmasked}}} \le 1 - \rho.
\end{equation}
The \methodname{} reward for rollout $a_i$ is
\begin{equation}
\mathcal{R}_{\text{\methodname}}(a_i) = -\sum_{v=1}^{T} m_v.
\label{eq:reward_sum}
\end{equation}
It is high when most reasoning tokens must be re-enabled to restore the answer probability. It penalizes tokens that can stay suppressed, i.e., tokens the answer does not need.

\begin{figure*}[t]
    \centering
    \includegraphics[width=\linewidth,trim={0.0cm 0 0.0cm 0},clip]{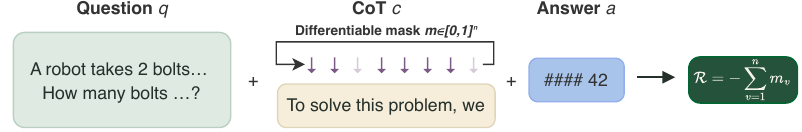}
    \caption{We initialize the additive attention mask $H^{\textsc{AtMan}}$ with 1 and multiply it by a negative constant. We then optimize the mask to restore the predicted answer probability. During the mask optimisation the values drop from 1 to 0 and remove the suppression of the reasoning tokens. More relevant tokens are unmasked earlier.}
    \label{fig:sparse_masking}
\end{figure*}
\subsection{Optimizing Saliency via Reinforcement Learning (RL)}
For RL, we combine the \textsc{\methodname} reward with a standard outcome reward:
\begin{equation}
\mathcal{R}_{\text{Outcome}}(a_i) =
\begin{cases}
0 & \text{if } p_i = j, \\
-1 & \text{otherwise},
\end{cases}
\end{equation}
where $j$ denotes the ground-truth answer and $p_i$ the rollout prediction of rollouts $a_i$.
Following GRPO \citep{shao2024deepseekmath} and GDPO \citep{liu2026gdpogrouprewarddecouplednormalization}, we compute the group-normalized reward mean $\mathcal{R}_{avg} = mean(\mathcal{R}(a_i))$ and reward standard deviation $\mathcal{R}_{std} = std(\mathcal{R}(a_i))$ across rewards to define the per-objective  advantage $\mathcal{A}(a_i)= \frac{\mathcal{R}(a_i)-\mathcal{R}_{avg}}{\mathcal{R}_{std}}$. Taking their weighted sum with objective specific weights  $w$  prevents reward scaling issues during zero-variance collapse:
\begin{flalign}
\label{eqn:gdpo}
\mathcal{A}_{total}(a_i) = \mathcal{A}_{\text{Accuracy}}(a_i) + \nonumber \\ 
\omega_{\text{\methodname}}\mathcal{A}_{\text{\methodname}}(a_i). 
\end{flalign}
Policy updates follow the clipped GRPO objective:

\begin{small}
\begin{flalign}
& \mathcal{L}_{\text{GRPO}}(\theta)=-\frac{1}{N}\sum\nolimits_{i=1}^{N}\frac{1}{|a_i|}\sum\nolimits_{t=1}^{|a_i|} \min\left(r_{i,t}\, \mathcal{A}_{total}(a_i),\right. \nonumber \\
& \left.\quad\; \text{clip}\!\left(r_{i,t},1-\epsilon,1+\epsilon\right) \mathcal{A}_{total}(a_i)\right), \nonumber\\
\end{flalign}
\end{small}
with
\begin{small}
\begin{flalign}
& r_{i,t}=\frac{\pi_\theta(a_{i,t} \mid q, a_{i,<t})}{\pi_{\theta_{\text{old}}}(a_{i,t} \mid q, a_{i,<t})}.
\end{flalign}
\end{small}

\textbf{Answer Format and Invalid Rollouts.}
We prompt the model to produce a reasoning trace followed by its final answer,
separated by a \texttt{\#\#\#\#} delimiter. The reasoning trace $c_{1:T}$
consists of all tokens generated before the delimiter, and the answer
$y_{1:N}$ of all tokens after it. To prevent reward hacking, we treat a rollout as invalid if (i)~the delimiter is missing, (ii)~it appears more than
once, (iii)~the reasoning trace is shorter than 15 tokens, or (iv)~the trace
repeats the prompt or copies its placeholders. Invalid rollouts receive the
outcome reward $-1$ and  in case of \textsc{\methodname} also $-max\_response\_length$.
For every valid rollout we compute the \textsc{AtManRL} reward,
\emph{regardless of whether the answer is correct}: we want to reward salient
reasoning even when it leads to a wrong answer, which keeps the saliency
signal informative in groups where all rollouts fail.

\section{Experimental Setup}
\label{sec:setup}
We train three instruction-tuned models in the 2--4B parameter range: Qwen2.5-3B-Instruct~\citep{qwen2.5}, Granite-3.3-2B-Instruct~\citep{granite2024}, and Phi-4-mini-instruct~\citep{microsoft2025phi4minitechnicalreportcompact}.
We train every model separately on each of four reasoning benchmarks: GSM8K~\citep{gsm8k_2021} (arithmetic problems), MMLU~\citep{hendryckstest2021} (multiple-choice questions), ProofWriter~\citep{sileo-2024-tasksource} (deductive reasoning over rule sets), and the GSM-E subset of QuestBench~\citep{li2025}.
One epoch consists of 7{,}500 training prompts, reshuffled every epoch; for datasets with fewer training prompts, we use all of them.
QuestBench's GSM-E subset consists of a single set of $\sim$6{,}500 samples, including several variants of each problem with different held-out variables, and has no evaluation split. We therefore generate a synthetic evaluation set (App.~\ref{sec:quest-eval}).

\textbf{Length-Penalty Baselines.}
We compare our method against a trade-off curve traced by a family of baselines LP($n$), trained from the same initial models.
The baselines use the same outcome reward but replace the \textsc{\methodname} reward with a length penalty, where $n \in \{30, 50, 100, 150, 300\}$ is the length budget.
The length reward enters the advantage exactly like the \textsc{\methodname} reward (Eq.~\ref{eqn:gdpo}), with weight $\omega = 0.1$.
We train \textsc{\methodname} at two auxiliary weights, $\omega \in \{0.1, 0.05\}$, on every model and dataset: $\omega = 0.1$ matches the baselines, and the lower weight yields longer traces, placing a second \textsc{\methodname} point along the length axis.

\textbf{Training.}
All runs use GRPO with GDPO advantages (Eq.~\ref{eqn:gdpo}); training parameters are listed in Appendix~\ref{sec:details}.
We train every run until accuracy and response length converge, sometimes with model- and dataset-specific learning rates, but always for the same number of epochs. Every setup is trained on a single seed.

\subsection{Evaluating Saliency and Faithfulness}
We evaluate our trained checkpoints on 1{,}000 held-out examples per dataset using greedy decoding.
We measure saliency with mistake injection (\S\ref{sec:eval-saliency}) and faithfulness with hint injection (\S\ref{sec:eval-faithfulness}).
We average each metric over (model, dataset) cells (\S\ref{sec:eval-protocol}) and plot it against the mean reasoning length of the original, unperturbed rollouts.
This allows us to compare \textsc{\methodname} with the LP($n$) trade-off curve at matched trace length.

\subsubsection{Mistake Injection}
\label{sec:eval-saliency}
We evaluate the models using mistake injection following \citet{lanham_measuring_2023}.
We generate a trace and answer, parse the trace into sentences, and let Qwen3.6-27B~\citep{qwen3.6-27b} insert errors.
At relative trace positions (0.1, 0.3, 0.5, 0.7, 0.9), we replace the original sentence with its corrupted version, let the model regenerate the remaining trace and answer, and measure how often the answer changes.
Externalizing the sentence manipulation eliminates self-generation bias, so that we measure trace saliency: the causal dependency of the final answer on preceding reasoning steps.

\subsubsection{Hint Injection}
\label{sec:eval-faithfulness}
Following \citet{chen2025reasoningmodelsdontsay}, we inject four hint types -- sycophancy, metadata, grader hacking, and unethical information -- into the prompt.
The hint is appended to the end of the prompt after a separator token and suggests an incorrect answer that differs from both the ground truth and the model's original response; for numerical datasets, hints are generated by adding or subtracting a small offset from the ground truth.
We then regenerate the CoT with the hint.
If the model switches its answer to the hinted one, it must have processed the hint internally; to remain faithful, it \emph{must} verbalize that hint.
We judge verbalization with Qwen3.6-27B~\citep{qwen3.6-27b} as an LLM judge, which makes a binary decision given the prompt and response.
Faithfulness is the probability that the model mentions the hint given that it switched from its original answer to the hinted answer, $P(\text{verbalization} \mid \text{switched to hint answer})$.

\subsubsection{Evaluation Protocol}
\label{sec:eval-protocol}
All metrics are computed per (model, dataset) cell and averaged over cells with equal weight, so that no single model or dataset dominates. The two evaluations differ in how samples are filtered.

\textbf{Mistake injection.}
We keep only samples where the judge confirms that the injected step is a mistake, the output is well formatted, and the mistake lands at its intended position.
Samples are pooled over the five insertion positions within each cell, which yields enough samples per cell for all 12 cells to enter the average.

\textbf{Hint injection.}
Faithfulness is estimated only on switches (\S\ref{sec:eval-faithfulness}) and pooled over hint types.
Cells with few switches are therefore noisy: at $p \approx 0.2$, $n = 20$ switches give a 95\% half-width of $\pm 17.5$\,pp ($E = z\sqrt{p(1-p)/n}$).
We require at least 20 judged switches per cell; if any method falls below this threshold, we drop the cell for all methods, so that every method is averaged over the same cells.
Trace length is averaged over the same cells, so the x-positions in the hint-injection figures differ from those in the mistake-injection figures.

\textbf{Comparison to the trade-off curve.}
We compare \textsc{\methodname} against the length-penalty trade-off curve, a linear interpolation through $LP$(30) to $LP$(300) ordered by mean trace length.
The gain of \textsc{\methodname} is its metric value minus the value of the curve at the same mean trace length. 95\% confidence intervals come from a paired, question-level bootstrap with 2{,}000 resamples: within each cell, evaluation questions are resampled jointly for all methods, and the cell matching for hint injection is re-applied within every resample.

\section{Experimental Results}
We first show that trace length alone trades saliency against faithfulness and use the LP($n$) baselines to trace this trade-off curve (\S\ref{sec:disentangle}).
We then compare \textsc{\methodname} against this curve in terms of saliency (\S\ref{sec:saliency}), faithfulness (\S\ref{sec:faithfulness}), and accuracy (\S\ref{sec:acc}), and finally study the convergence ratio $\rho$ (\S\ref{sec:rho}) and the computational overhead (\S\ref{sec:overhead}).

\subsection{Length, Saliency, and Faithfulness}
\label{sec:disentangle}
The length-penalty baselines LP($n$) show that saliency and faithfulness both depend on trace length, and in opposite directions (Fig.~\ref{fig:mistake_main} and Fig.~\ref{fig:hint_main}).
Under mistake injection, the probability that an inserted mistake changes the final answer decreases from about 59\% for LP(50) to about 46\% for LP(300); LP(30) lies slightly below LP(50), and the interpolated curve passes through both.
Under hint injection, faithfulness increases from about 6\% for LP(30) to about 29\% for LP(300).
Both trends are consistent with how the two metrics are defined.
In a short trace, each remaining step carries more of the computation, so corrupting one is more likely to change the answer.
A longer trace has more room to make explicit the factors that influence the answer, including an injected hint.
Shortening traces with a length penalty therefore increases saliency at the cost of faithfulness, and comparing methods that produce traces of different lengths is confounded by length alone.
Therefore, in what follows, we compare \textsc{\methodname} not against a single baseline but against the trade-off curve traced by the LP($n$) baselines (\S\ref{sec:eval-protocol}).

\subsection{Saliency}
\label{sec:saliency}
\begin{figure}[t]
    \centering
    \includegraphics[width=0.99\linewidth, trim=0 0.2cm 0 0, clip]{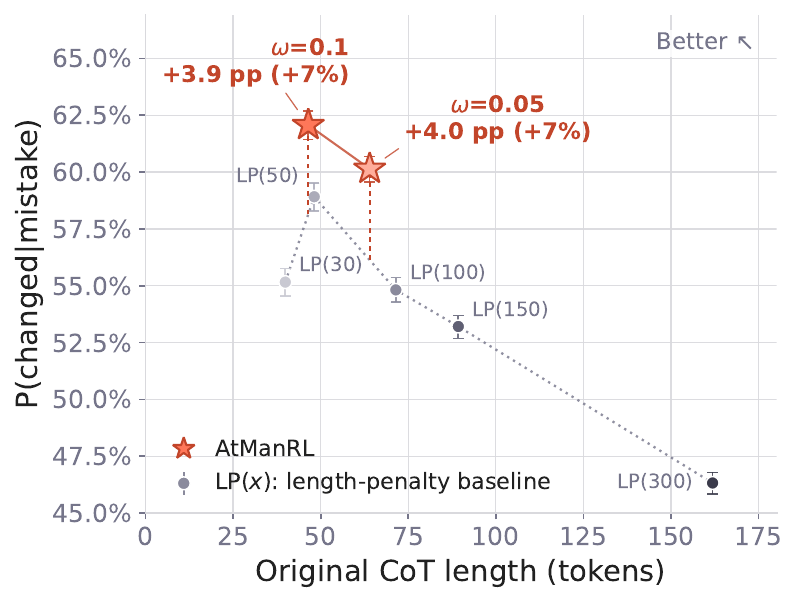}
    \caption{\textbf{Saliency:} Probability of a changed answer given an inserted mistake, plotted over the mean CoT length and averaged over all models and datasets. Dotted line: LP($n$) trade-off curve; error bars: 95\% paired bootstrap confidence intervals (\S\ref{sec:eval-protocol}).}
    \label{fig:mistake_main}
\end{figure}

Fig.~\ref{fig:mistake_main} shows the probability that the final answer changes when we insert a mistake into the reasoning trace and let the model continue from it.
A model whose answer changes more often after an inserted mistake produces more salient reasoning traces, i.e., traces on which the answer causally depends.
Across the LP($n$) baselines, this probability decreases with trace length, which yields a clear trade-off curve across three models and four datasets.
\textsc{\methodname} lies above this curve by 3.9\,pp ($\omega = 0.1$) and 4.0\,pp ($\omega = 0.05$) at matched trace length; the 95\% bootstrap confidence intervals are shown in Fig.~\ref{fig:mistake_main}.

\begin{figure}[t]
    \centering
    \includegraphics[width=0.99\linewidth, trim=0 0.2cm 0 0, clip]{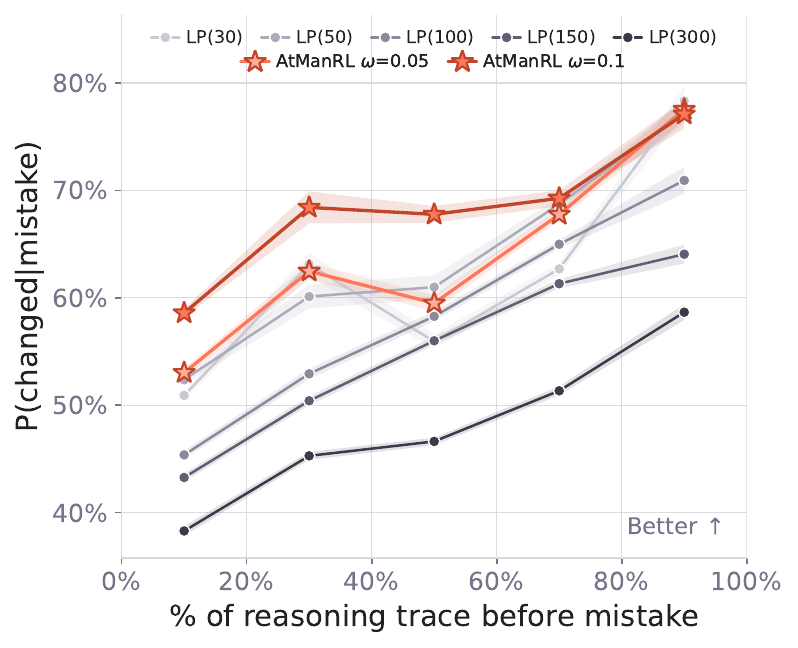}
    \caption{\textbf{Saliency:} Probability of a changed answer given an inserted mistake, plotted over the relative position in the reasoning trace at which the mistake is inserted and averaged over all models and datasets, following \citet{lanham_measuring_2023}. Shaded bands show $\pm1$ binomial standard error.}
    \label{fig:mistake_lanham}
\end{figure}

Fig.~\ref{fig:mistake_lanham} breaks this result down by insertion position, following the presentation of \citet{lanham_measuring_2023}.
The probability of a changed answer rises the later the mistake is inserted, suggesting that later parts of the reasoning trace have a larger causal influence on the final answer, although later mistakes also leave the model less room to recover.
\textsc{\methodname} lies above the baselines of similar length at almost every insertion point.

\subsection{Faithfulness}
\label{sec:faithfulness}
\begin{figure}[t]
    \centering
    \includegraphics[width=0.99\linewidth, trim=0 0.2cm 0 0, clip]{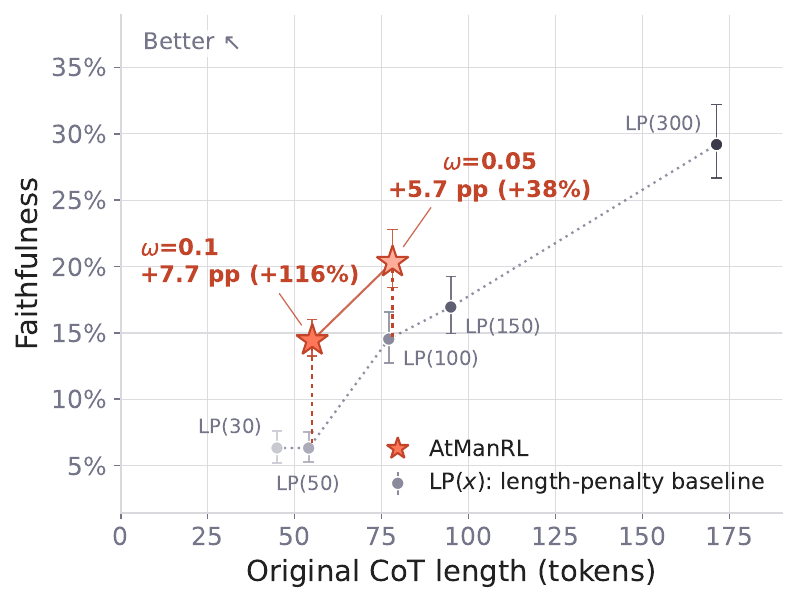}
    \caption{\textbf{Faithfulness} under hint injection, plotted over the mean CoT length and averaged over hint types and the (model, dataset) cells that pass the switch filter. \textsc{\methodname} is more faithful than baselines of similar CoT length. Dotted line: LP($n$) trade-off curve; error bars: 95\% paired bootstrap confidence intervals (\S\ref{sec:eval-protocol}).}
    \label{fig:hint_main}
\end{figure}

Fig.~\ref{fig:hint_main} shows that the saliency reward of \textsc{\methodname} also improves faithfulness.
Along the baseline curve, the relationship with length is reversed: the longer the reasoning, the more faithful.
This follows from how faithfulness is measured.
Hint injection only scores traces in which the hint changed the final answer, and \citet{chen2025reasoningmodelsdontsay} argue that a model that switches to the hinted answer must have used the hint and therefore has to verbalize it to be faithful.
Pure length pressure leaves less room in the trace to do so, so faithfulness drops for shorter traces.
Nevertheless, \textsc{\methodname} lies above the length-matched curve by 7.7\,pp ($\omega = 0.1$) and 5.7\,pp ($\omega = 0.05$), relative increases of 116\% and 38\%, averaged over four hint types and the 9 of 12 cells that pass the switch filter (\S\ref{sec:eval-protocol}); the 95\% bootstrap confidence intervals are shown in Fig.~\ref{fig:hint_main}. The large relative gain at $\omega = 0.1$ reflects the low faithfulness of the curve at short trace lengths (about 7\%), so the absolute differences are the more informative comparison.
Across individual models and datasets (App.~\ref{app:hint}), \textsc{\methodname} lies on or above the curve in every breakdown at $\omega = 0.05$, and falls below it in two of seven at $\omega = 0.1$ (QuestBench, $-0.2$\,pp; ProofWriter, where hints rarely change the answer, $-2.4$\,pp).

\subsection{Accuracy}
\label{sec:acc}
\begin{figure}[t]
    \centering
    \includegraphics[width=0.99\linewidth, trim=0 0.3cm 0 0, clip]{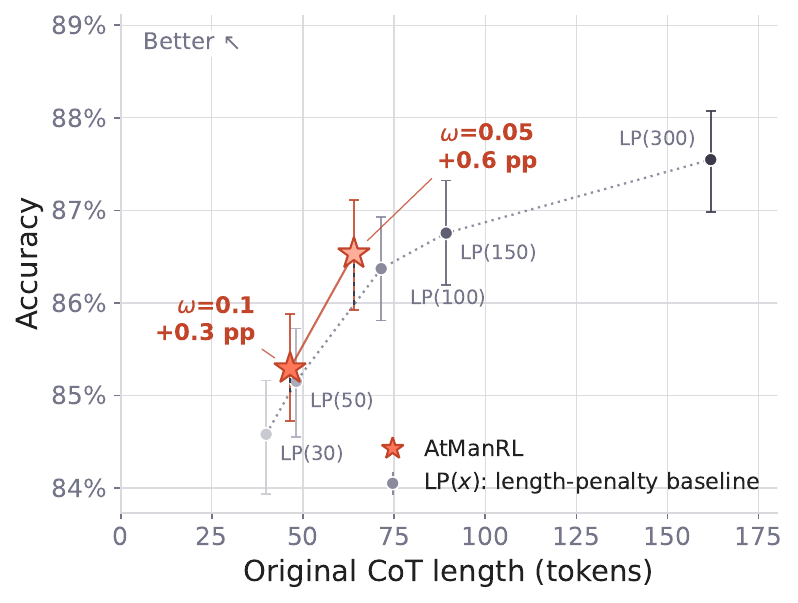}
    \caption{\textbf{Accuracy} plotted over the mean CoT length and averaged over all models and datasets. Dotted line: LP($n$) trade-off curve; error bars: 95\% paired bootstrap confidence intervals (\S\ref{sec:eval-protocol}).}
    \label{fig:acc_main}
\end{figure}

A higher saliency or faithfulness score would be of little use if it came at the cost of task performance.
Fig.~\ref{fig:acc_main} therefore shows the accuracy of all runs over their mean reasoning length, again averaged over the three models and four datasets.
The accuracy of \textsc{\methodname} matches the LP($n$) curve: the differences of +0.3\,pp ($\omega = 0.1$) and +0.6\,pp ($\omega = 0.05$) lie within the confidence intervals (Fig.~\ref{fig:acc_main}), so \textsc{\methodname} neither improves nor degrades accuracy relative to its length-matched baselines.
The gains in saliency (\S\ref{sec:saliency}) and faithfulness (\S\ref{sec:faithfulness}) therefore do not come from trading off correctness.

\subsection{$\rho$ Ablation}
\label{sec:rho}
\begin{figure}[t]
    \centering
    \includegraphics[width=0.99\linewidth, trim=0 0.4cm 0 0, clip]{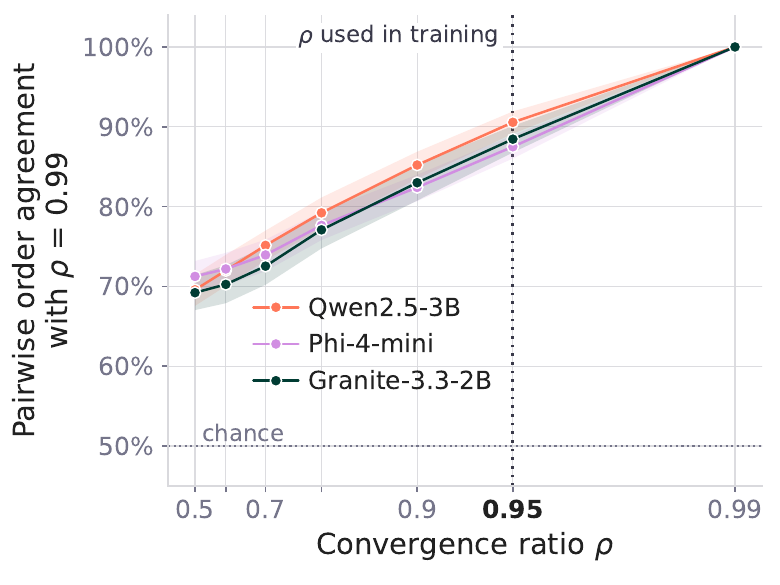}
    \caption{Pairwise agreement of in-group rollout rankings by the \textsc{\methodname} reward, relative to $\rho = 0.99$, using the mask learning rate from training. The training value reaches $\sim$90\% agreement.}
    \label{fig:rho_pairwise}
\end{figure}
The convergence ratio $\rho$ determines when the mask optimization stops and therefore how the \textsc{\methodname} reward is constructed.
In training, we use $\rho = 0.95$.
Fig.~\ref{fig:rho_pairwise} shows that this value achieves 90\% pairwise agreement with $\rho = 0.99$, where agreement is computed over the in-group rankings of rollouts by the \textsc{\methodname} reward.
We evaluate 100 groups using the mask learning rate and step cap from training, and use the mask at the last optimization step for the comparison.

\subsection{Overhead Evaluation}
\label{sec:overhead}
\begin{figure}[t]
    \centering
    \includegraphics[width=0.99\linewidth, trim=0 0.4cm 0 0, clip]{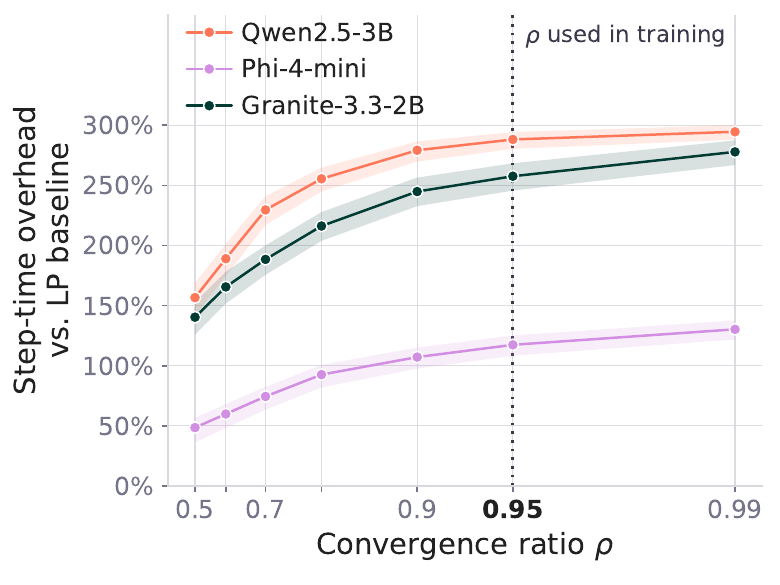}
    \caption{Step-time overhead of \textsc{\methodname} relative to the LP($n$) baseline for different convergence ratios $\rho$.}
    \label{fig:overhead}
\end{figure}
We apply the \textsc{\methodname} mask in every layer, its overhead depends on the architecture, the number of layers, and the response length; it is therefore largest when responses are longest, typically early in training.
We use the standard SDPA attention implementation and are working on a more efficient variant.
Fig.~\ref{fig:overhead} shows that the overhead grows with $\rho$ and depends strongly on the model: at $\rho = 0.95$, it lies between 110\% and 280\% relative to the baseline.
While substantial, we consider this overhead acceptable for a post-training method.

\section{Conclusions}
We introduced \textsc{AtManRL}, a method that trains LLMs to produce salient reasoning through differentiable attention manipulation. By optimizing an attention mask that identifies the reasoning tokens needed to recover the model's answer, we derive a saliency reward that we combine with an outcome reward during RL. Since saliency and faithfulness both depend on trace length, and in opposite directions, we evaluate against a trade-off curve of length-penalty baselines instead of a single run. Across three models, four datasets, and two \textsc{\methodname} weight settings, \textsc{\methodname} lies above this curve by $3.9-4.0$pp under mistake injection and by $5.7-7.7$pp under hint injection, while matching the accuracy when compared to similar length baselines. These results suggest that training explicitly for saliency is a promising route towards CoT reasoning that is not only more salient but also more faithful to the model's actual decision process.

\section{Limitations} \label{sec:limitations}
First, all evaluated models are in the 2–4B parameter range on a single seed only. It remains open whether the saliency dynamics we observe and the reward's effect on accuracy and trace length hold at larger scale, where reasoning traces and attention patterns may behave differently.
Furthermore, our current reward structure prioritizes information density and conciseness, which may inadvertently discourage the backtracking and self-correction behaviours recently highlighted in the community. Current discussions -- particularly following SDPO (Self-Distillation Policy Optimization) paper \citep{hubotter2026reinforcementlearningselfdistillation}  -- suggest that effective reasoning often requires dense credit assignment and the ability to retrospectively identify and fix mistakes \citep{kim2026doesselfdistillationsometimesdegrade}. While \textsc{\methodname} removes non-salient tokens, it may also suppress tokens that signify a model’s transition from a wrong path to a right one (e.g., "Wait, let me rethink"). While we did not perform a detailed analysis of this phenomenon yet, we hypothesize that future refinements to the reward function could explicitly protect these self-correction markers to ensure that efficiency does not come at the cost of reasoning depth.
We are also working on efficiency of the method. That means we are looking into more efficient attention implementations and are planning to ablate over the actual layers that are needed to achieve the desired effect, which could also drastically reduce the overhead.

\section{Ethical Considerations}

This work studies whether chain-of-thought reasoning causally supports a model’s answer, and proposes a training objective that strengthens this dependence. Saliency, as we measure it, is one property a reasoning trace can have. Faithfulness is harder to measure directly, so we operationalize it through hint verbalization \cite{chua2025deepseekr1reasoningmodels, chen2025reasoningmodelsdontsay} as a proxy for whether a trace reports the factors that drove the answer. Improvements on these proxies should not be read as evidence that a model is honest, safe, or trustworthy, and traces produced by \textsc{\methodname} should not be used as the sole basis for judging a model’s decision process.

Faithfulness of chain-of-thought explanations is a broader open problem for LLMs \cite{barez2025chain} -- with LLM faithfulness being generally low, and our results reflect this: even with \textsc{\methodname}, the verbalization rate averages roughly 14–21\% in our evaluations, which is far from sufficient to rely on traces as an explanation. We view this as a general concern for reasoning traces rather than a specific failure of our method. Within that setting, \textsc{\methodname} moves both metrics in the desired direction at matched trace length and without accuracy loss. We see this as a step towards more faithful reasoning, and losing the gap will require substantial work across the field.

Our evaluation relies on a single LLM judge (Qwen3.6-27B) both to filter invalid mistakes and to decide whether a hint was verbalized. Such judges may carry systematic biases, and we did not conduct a human agreement study to quantify them. Since the judge applies the same criteria to every method, we expect these biases to affect the comparisons between methods only to a limited extent. We therefore focus on relative differences rather than absolute values.

\bibliography{bib}

\clearpage
\appendix
\section{Appendix}
\subsection{Questbench Evaluation}
\label{sec:quest-eval}
We generate a synthetic evaluation set of Questbench's GSM-E subset because there is no separate evaluation split for it. For each problem we resample every constant (integers greater than one uniformly from $[\lfloor x/2 \rfloor, 2x]$, decimals from $[x/2, 2x]$ at their original precision and recompute all intermediate values and the answer by executing the problem's equations. We keep an instance only if every intermediate value retains its original sign and integrality, the answer is an integer, and the answer differs from every training answer for the same problem. Underspecified variants are QuestBench's own rewritten problems with the same new constants. We verify every item by solving its equation system exactly with SymPy: complete problems must have a unique solution equal to the label, and underspecified problems must leave the answer undetermined, while adding back only the held-out variable must recover it. The final set contains 1{,}000 items (500 complete, 500 underspecified) drawn from 773 distinct problems, and no item text or answer occurs in the training data. The equation templates themselves are shared with training, so this set measures whether models compute correct answers for familiar problem structures with unseen values, rather than generalization to new problems.

\onecolumn
\subsection{Training Details}
\label{sec:details}
Training details, hyper parameters, and settings are in Tables~\ref{tab:train-shared-final}, ~\ref{tab:train-runs-final}, ~\ref{tab:train-atman-final}

\begin{table*}[h]
\centering
\small
\begin{tabularx}{\linewidth}{@{}lX@{}}
\toprule
Setting & Value \\
\midrule
\multicolumn{2}{l}{\emph{Policy optimization}} \\
Algorithm & GRPO with GDPO advantages: each reward channel is $z$-scored within its group (zero if the group has no variance), then weighted and summed \\
Advantage & $A=\tilde r_{\mathrm{corr}}+w\,\tilde r_{\mathrm{aux}}$, $w=0.1$ for both arms ($r_{\mathrm{aux}}=r_{\mathrm{len}}$ for LP($n$), $r_{\mathrm{AtMan}}$ for \textsc{\methodname}), except the sub-4B \textsc{\methodname} weight, $\omega=0.05$ \\
Surrogate loss & PPO clip, token-level ratio, $\epsilon=0.2$; token mean per sequence, then mean over sequences \\
KL penalty & none ($\beta=0$, no reference model) \\
Passes per step & 1 (one optimizer step per mini-batch, single pass over the step's rollouts) \\
Optimizer & AdamW, $\beta=(0.9,0.999)$, $\epsilon=10^{-8}$, weight decay 0 \\
Gradient clipping & global norm 1 \\
LR schedule & constant, no warmup; continuation legs resume weights and optimizer state, mostly at half the LR (per-run values in Table~\ref{tab:train-runs-final}) \\
\midrule\multicolumn{2}{l}{\emph{Batching and compute}} \\
GPUs / parallelism & 48, data-parallel: DDP for the sub-4B models (no ZeRO/FSDP); activation checkpointing \\
Precision & bf16 weights, gradients and AdamW moments (no fp32 master copy) \\
Prompts per step & 384 (8/GPU) \\
Group size $G$ & 8 rollouts per prompt (3,072 rollouts per step) \\
Policy mini-batch & 16 rollouts/GPU (768 global), i.e.\ 4 updates per step \\
\midrule\multicolumn{2}{l}{\emph{Data and rollouts}} \\
Prompts per epoch & 7,500: the first 7,500 items of the train split, cycled if shorter, reshuffled every epoch (20 steps) \\
Train splits & GSM8K train (7,473); MMLU auxiliary\_train (99,842); ProofWriter train (585,552); QuestBench GSM-E train (5,272) \\
Max prompt / response & 512 / 512 tokens \\
Rollout sampling & temperature 1.0, top-$p$ 0.95, top-$k$ 50 \\
\midrule\multicolumn{2}{l}{\emph{Rewards}} \\
Correctness & $r_{\mathrm{corr}}=0$ if the answer after \texttt{\#\#\#\#} is correct, $-1$ otherwise \\
Format & no separate reward; no \texttt{\#\#\#\#}, \texttt{\#\#\#\#} inside the CoT, a CoT under 15 tokens or copied prompt placeholders count as incorrect (and get the worst AtMan score) \\
Length penalty LP($n$) & $r_{\mathrm{len}}=-0.1\min(1,L/500)\,\mathbf{1}[L>n]$ on correct rollouts, 0 otherwise; $L$ = CoT tokens before \texttt{\#\#\#\#}; $n\in\{30, 50, 100, 150, 300\}$\\
\textsc{\methodname} & $r_{\mathrm{AtMan}}=-\sum_t m_t$ over CoT tokens (Table~\ref{tab:train-atman-final}), $\omega=0.1$ and $\omega = 0.05$ in the sub-4B); no length penalty \\
Seed & 42 (one run per arm) \\
\bottomrule
\end{tabularx}
\caption{Training hyperparameters shared by every run in the hint-injection and mistake-continuation figures (Qwen2.5-3B, Phi-4-mini, Granite-3.3-2B on GSM8K, MMLU, ProofWriter, QuestBench;\textsc{\methodname} and LP($n$); and the sub-4B \textsc{\methodname} runs at AtManSum weight $w=0.05$, which differ from the $w=0.1$ arm only in the weight). Where a setting differs between models, its row names the exception. Per-run schedules are in Table~\ref{tab:train-runs-final}.}
\label{tab:train-shared-final}
\end{table*}

\begin{table*}[h]
\centering
\resizebox{\linewidth}{!}{%
\begin{tabular}{@{}lllllrl@{}}
\toprule
Model & Dataset & Arm & Schedule: epochs\,@\,LR ($10^{-6}$) & Ep. & Steps & AtMan $T_{\max}$ / $\eta_m$ / $B_m$ \\
\midrule
Qwen2.5-3B & GSM8K & \textsc{\methodname} ($w{=}0.1$) & 5\,@\,10 + 1\,@\,5 & 6 & 120 & 100 / 1 / 2 \\
 &  & \textsc{\methodname} ($w{=}0.05$) & 5\,@\,10 + 1\,@\,5 & 6 & 120 & 100 / 1 / 2 \\
 &  & LP($n$), $n\in\{30,50,100,150,300\}$ & 5\,@\,10 + 1\,@\,5 & 6 & 120 & -- \\
\cmidrule(l){2-7}
 & MMLU & \textsc{\methodname} ($w{=}0.1$) & 5\,@\,16 + 2\,@\,8 & 7 & 140 & 100 / 1 / 1 \\
 &  & \textsc{\methodname} ($w{=}0.05$) & 5\,@\,16 + 2\,@\,8 & 7 & 140 & 100 / 1 / 1 \\
 &  & LP($n$), $n\in\{30,50,100,150,300\}$$^{\ddagger}$ & 3\,@\,16 + 4\,@\,8 & 7 & 140 & -- \\
\cmidrule(l){2-7}
 & ProofWriter & \textsc{\methodname} ($w{=}0.1$) & 5\,@\,12 & 5 & 100 & 75 / 0.1 / 1 \\
 &  & \textsc{\methodname} ($w{=}0.05$) & 5\,@\,12 & 5 & 100 & 75 / 0.1 / 1 \\
 &  & LP($n$), $n\in\{30,50,100,150,300\}$ & 5\,@\,12 & 5 & 100 & -- \\
\cmidrule(l){2-7}
 & QuestBench & \textsc{\methodname} ($w{=}0.1$) & 5\,@\,4 & 5 & 100 & 100 / 1 / 2 \\
 &  & \textsc{\methodname} ($w{=}0.05$) & 5\,@\,4 & 5 & 100 & 100 / 1 / 2 \\
 &  & LP($n$), $n\in\{30,50,100,150,300\}$$^{\ddagger}$ & 3\,@\,4 + 2\,@\,2 & 5 & 100 & -- \\
\midrule
Phi-4-mini & GSM8K & \textsc{\methodname} ($w{=}0.1$) & 5\,@\,4 + 1\,@\,2 & 6 & 120 & 75 / 1 / 4 \\
 &  & \textsc{\methodname} ($w{=}0.05$) & 5\,@\,4 + 1\,@\,2 & 6 & 120 & 75 / 1 / 4 \\
 &  & LP($n$), $n\in\{30,50,100,150,300\}$ & 5\,@\,4 + 1\,@\,2 & 6 & 120 & -- \\
\cmidrule(l){2-7}
 & MMLU & \textsc{\methodname} ($w{=}0.1$) & 5\,@\,8 + 2\,@\,4 & 7 & 140 & 75 / 1 / 1 \\
 &  & \textsc{\methodname} ($w{=}0.05$) & 5\,@\,8 + 2\,@\,4 & 7 & 140 & 75 / 1 / 1 \\
 &  & LP($n$), $n\in\{30,50,100,150,300\}$$^{\ddagger}$ & 3\,@\,8 + 4\,@\,4 & 7 & 140 & -- \\
\cmidrule(l){2-7}
 & ProofWriter & \textsc{\methodname} ($w{=}0.1$) & 5\,@\,8 & 5 & 100 & 75 / 1 / 2 \\
 &  & \textsc{\methodname} ($w{=}0.05$) & 5\,@\,8 & 5 & 100 & 75 / 1 / 2 \\
 &  & LP($n$), $n\in\{30,50,100,150,300\}$ & 5\,@\,8 & 5 & 100 & -- \\
\cmidrule(l){2-7}
 & QuestBench & \textsc{\methodname} ($w{=}0.1$) & 5\,@\,8 & 5 & 100 & 75 / 0.1 / 4 \\
 &  & \textsc{\methodname} ($w{=}0.05$) & 5\,@\,8 & 5 & 100 & 75 / 0.1 / 4 \\
 &  & LP($n$), $n\in\{30,50,100,150,300\}$ & 5\,@\,8 & 5 & 100 & -- \\
\midrule
Granite-3.3-2B & GSM8K & \textsc{\methodname} ($w{=}0.1$) & 5\,@\,2 + 1\,@\,1 & 6 & 120 & 100 / 1 / 1 \\
 &  & \textsc{\methodname} ($w{=}0.05$) & 5\,@\,2 + 1\,@\,1 & 6 & 120 & 100 / 1 / 1 \\
 &  & LP($n$), $n\in\{30,50,100,150,300\}$ & 5\,@\,2 + 1\,@\,1 & 6 & 120 & -- \\
\cmidrule(l){2-7}
 & MMLU & \textsc{\methodname} ($w{=}0.1$) & 5\,@\,4 + 2\,@\,2 & 7 & 140 & 100 / 1 / 1 \\
 &  & \textsc{\methodname} ($w{=}0.05$) & 5\,@\,4 + 2\,@\,2 & 7 & 140 & 100 / 1 / 1 \\
 &  & LP($n$), $n\in\{30,50,100,150,300\}$$^{\ddagger}$ & 3\,@\,4 + 4\,@\,2 & 7 & 140 & -- \\
\cmidrule(l){2-7}
 & ProofWriter & \textsc{\methodname} ($w{=}0.1$) & 5\,@\,4 & 5 & 100 & 75 / 1 / 4 \\
 &  & \textsc{\methodname} ($w{=}0.05$) & 5\,@\,4 & 5 & 100 & 75 / 1 / 4 \\
 &  & LP($n$), $n\in\{30,50,100,150,300\}$ & 5\,@\,4 & 5 & 100 & -- \\
\cmidrule(l){2-7}
 & QuestBench & \textsc{\methodname} ($w{=}0.1$) & 5\,@\,4 & 5 & 100 & 75 / 1 / 4 \\
 &  & \textsc{\methodname} ($w{=}0.05$) & 5\,@\,4 & 5 & 100 & 75 / 1 / 4 \\
 &  & LP($n$), $n\in\{30,50,100,150,300\}$ & 5\,@\,4 & 5 & 100 & -- \\
\bottomrule
\end{tabular}}
\caption{Per-run training schedule of every checkpoint in the hint-injection and mistake-continuation figures (Qwen2.5-3B, Phi-4-mini, Granite-3.3-2B on GSM8K, MMLU, ProofWriter, QuestBench), including the sub-4B AtManSum-weight ($\omega=0.1$ and $\omega=0.05$). A schedule lists the epochs trained at each learning rate; a later entry is a continuation leg resumed from the previous checkpoint (weights and optimizer state). One epoch is 7,500 prompts; a step is one batch of rollouts. $T_{\max}$: AtMan step cap; $\eta_m$: mask learning rate; $B_m$: AtMan mini-batch. $^{\ddagger}$Same length as \textsc{\methodname}, different split between the two learning rates.}
\label{tab:train-runs-final}
\end{table*}

\begin{table}[h]
\centering
\small
\begin{tabularx}{\linewidth}{@{}l>{\raggedright\arraybackslash}X@{}}
\toprule
Setting & Value \\
\midrule
Reward & $r_{\mathrm{AtMan}}=-\sum_{t\in\mathrm{CoT}} m_t$ at the stopping step; advantage weight for all arms $\omega=0.1$ but the \textsc{\methodname} second weight settinga: $\omega=0.05$\\
Mask & one suppression weight $m_t\in[0,1]$ per CoT token, initialised at 1 (fully suppressed) \\
Suppression scale $c$ & -10 \\
Mask objective & cross-entropy of the policy's own answer tokens after \texttt{\#\#\#\#}, content tokens only (digits, option letters) \\
Fidelity stop & first step with $\mathrm{gap}(t)/\mathrm{gap}(0)\le 1-\rho$, $\rho=0.95$; gap = answer CE increase over the unmasked model; the mask at $T_{\max}$ if never reached; $m\equiv1$ if $\mathrm{gap}(0)<0.05$ nats \\
Mask optimizer & AdamW, $\beta=(0.9,0.9999)$, weight decay 0.05; LR $\eta_m$ decayed linearly to $10^{-3}\eta_m$ over $T_{\max}$ steps \\
Per-run settings & $T_{\max}\in\{75, 100\}$, $\eta_m\in\{0.1, 1\}$, $B_m\in\{1, 2, 4\}$ (Table~\ref{tab:train-runs-final}) \\
Scored rollouts & every rollout with a valid format, correct or not \\
\bottomrule
\end{tabularx}
\caption{AtMan attribution settings of the \textsc{\methodname} arms.}
\label{tab:train-atman-final}
\end{table}

\newpage
\twocolumn
\subsection{Token-Category Analysis}
\begin{figure}[H]
    \centering
    \includegraphics[width=0.99\linewidth]{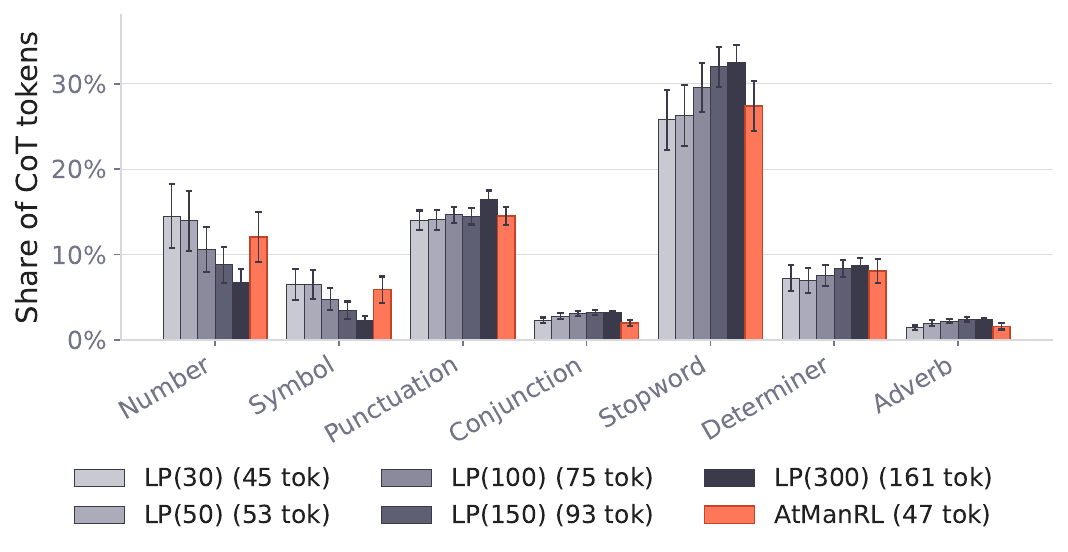}
    \caption{Statistical Analysis of word distribution}
    \label{fig:stats_analysis}
\end{figure}

To see whether \textsc{\textsc{\methodname}} ($\omega = 0.1$) changes \emph{what} the model writes, we classify all CoT tokens into lexical categories and compare
their types across runs (Fig.~\ref{fig:stats_analysis}). Since the mask removes tokens that do not contribute to the answer, we expected \textsc{\textsc{\methodname}} to produce the lowest share of stop words. Instead, compared to the length-matched baselines LP(30) and LP(50), \textsc{\textsc{\methodname}} traces contain slightly more stop words and fewer numbers and symbols, although most differences lie within the error
bars. This suggests that \textsc{\textsc{\methodname}} does not only compress traces the way a length penalty does: at comparable length, it keeps connective tokens that a pure length penalty removes, possibly because they structure the reasoning that the answer depends on.

\onecolumn
\label{sec:stats}
\twocolumn

\newpage
\section{Hint injection per-model and per-dataset}
\label{app:hint}
\begin{figure}[H]
    \centering
    \includegraphics[width=0.99\linewidth, trim=0 0.2cm 0 0, clip]{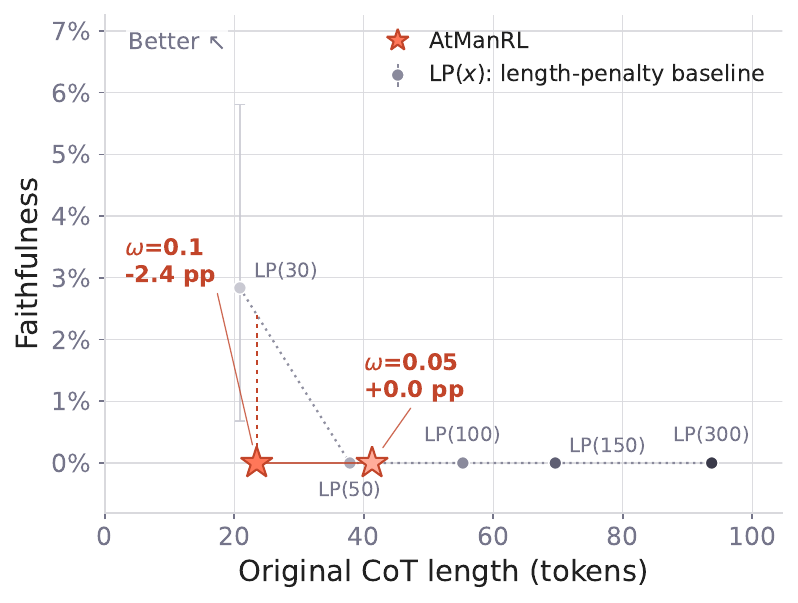}
    \caption{Hint injection on ProofWriter across sub-4B models (greedy decoding): CoT length and faithfulness.}
    \label{fig:hint_dataset_proofwriter}
\end{figure}

To complement the hint injection results in the main text we also provide faithfulness plots for each model and datasets. When averaging over the models or datasets we average them all equally. We can also see in Fig.~\ref{fig:hint_dataset_proofwriter} that the Proofwriter dataset has a very low faithfulness rate. This is due to the low due switch rate of this dataset and very short trace length. That means if a hint is inserted it is very rarely used.
\subsection{Per-dataset}
\begin{figure}[H]
    \centering
    \includegraphics[width=0.99\linewidth, trim=0 0.2cm 0 0, clip]{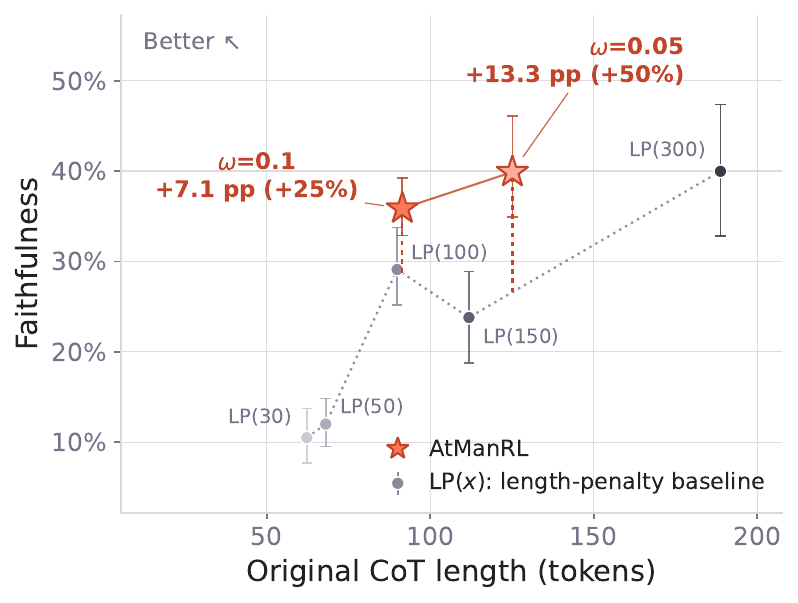}
    \caption{Hint injection on GSM8K across sub-4B models (greedy decoding): CoT length and faithfulness.}
    \label{fig:hint_dataset_gsm8k}
\end{figure}
\begin{figure}[H]
    \centering
    \includegraphics[width=0.99\linewidth, trim=0 0.2cm 0 0, clip]{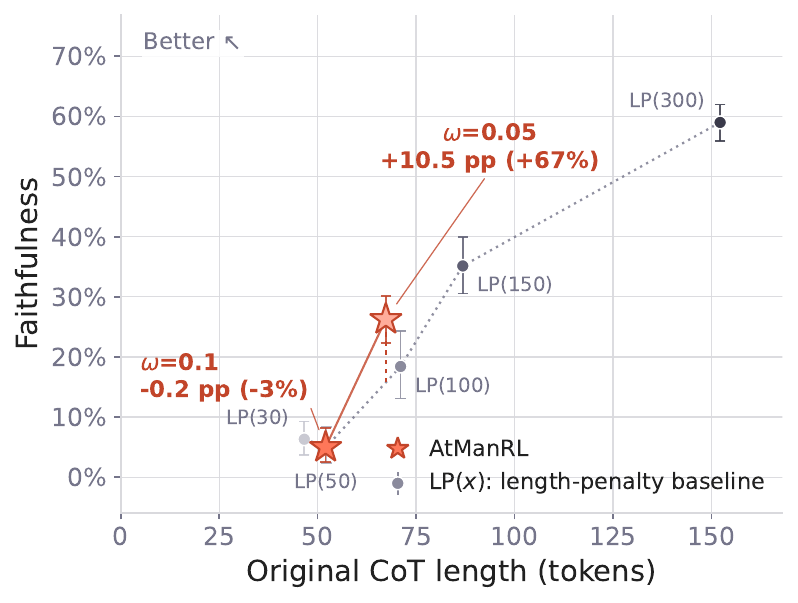}
    \caption{Hint injection on QuestBench across sub-4B models (greedy decoding): CoT length and faithfulness.}
    \label{fig:hint_dataset_questbench}
\end{figure}
\begin{figure}[H]
    \centering
    \includegraphics[width=0.99\linewidth, trim=0 0.2cm 0 0, clip]{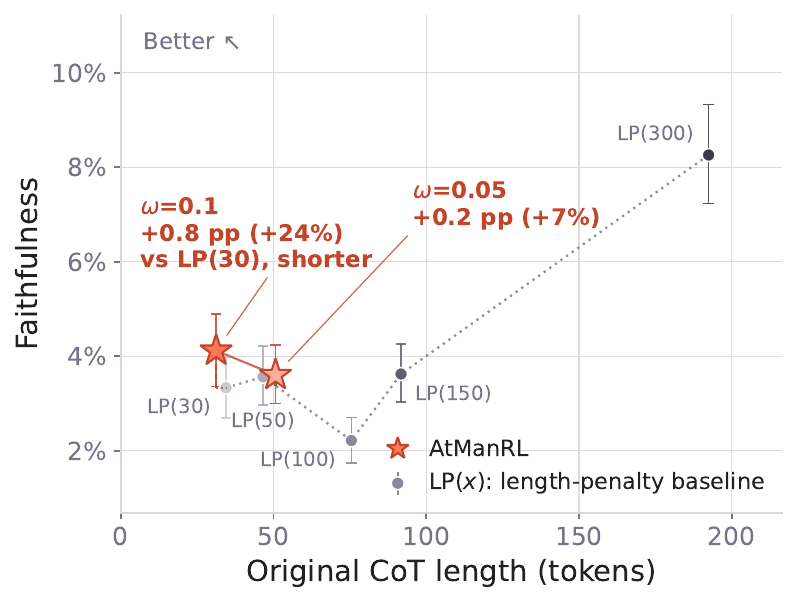}
    \caption{Hint injection on MMLU across sub-4B models (greedy decoding): CoT length and faithfulness.}
    \label{fig:hint_dataset_mmlu}
\end{figure}
\subsection{Per-model}
\begin{figure}[H]
    \centering
    \includegraphics[width=0.99\linewidth, trim=0 0.2cm 0 0, clip]{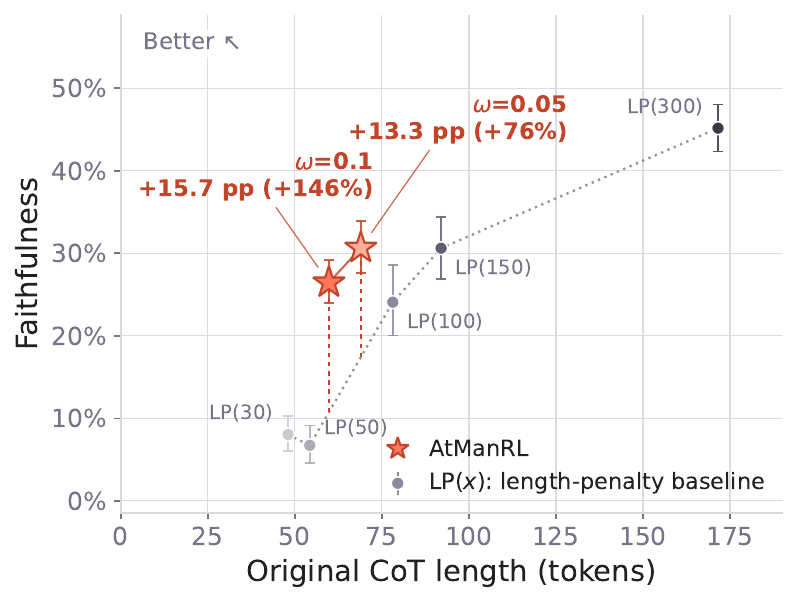}
    \caption{Hint injection on Qwen2.5-3B, pooled across datasets (greedy decoding): CoT length and faithfulness.}
    \label{fig:hint_model_qwen25_3b}
\end{figure}
\begin{figure}[H]
    \centering
    \includegraphics[width=0.99\linewidth, trim=0 0.2cm 0 0, clip]{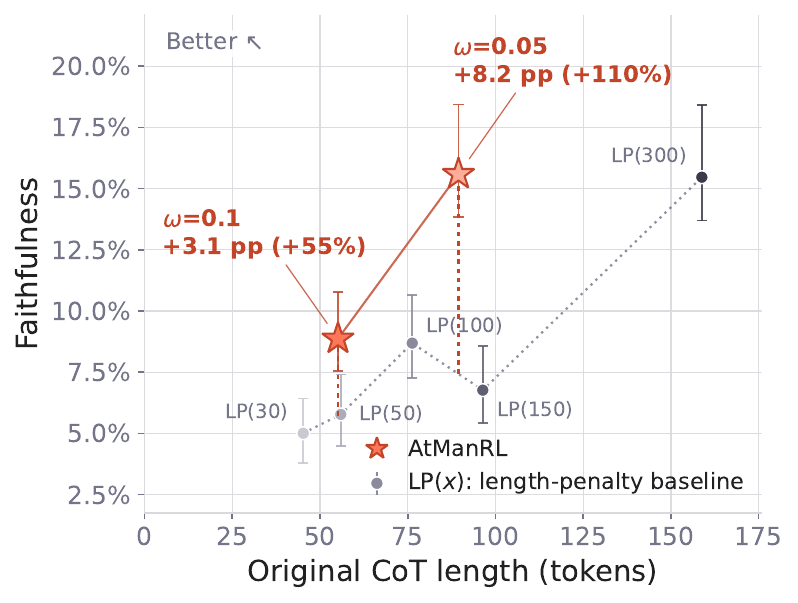}
    \caption{Hint injection on Granite-3.3-2B, pooled across datasets (greedy decoding): CoT length and faithfulness.}
    \label{fig:hint_model_granite33_2b}
\end{figure}
\begin{figure}[H]
    \centering
    \includegraphics[width=0.99\linewidth, trim=0 0.2cm 0 0, clip]{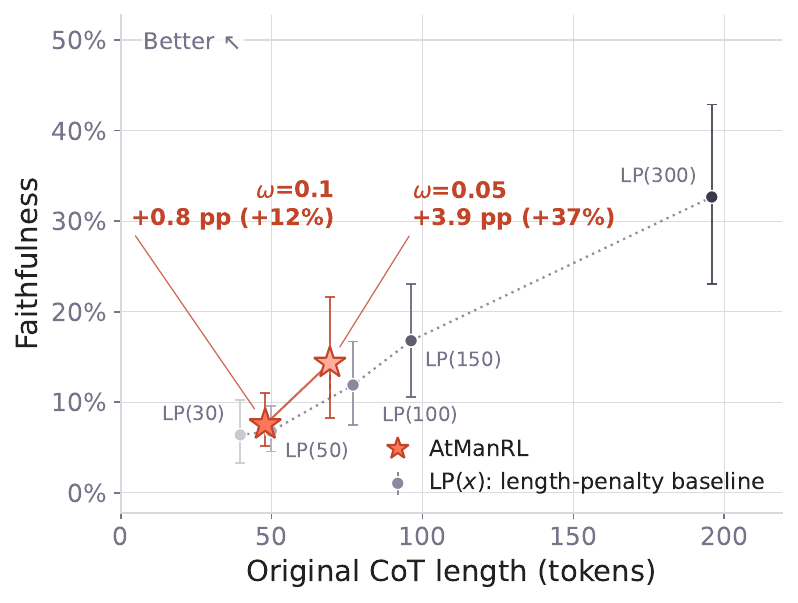}
    \caption{Hint injection on Phi-4-mini-3.8B, pooled across datasets (greedy decoding): CoT length and faithfulness.}
    \label{fig:hint_model_phi4mini}
\end{figure}

\section{Mistake injection per-model and per-dataset}
To complement the mistake injection results in the main text we also provide saliency plots for each model and datasets. When averaging over the models or datasets we average them all equally.
\subsection{Per-dataset}
\begin{figure}[H]
    \centering
    \includegraphics[width=0.99\linewidth, trim=0 0.2cm 0 0, clip]{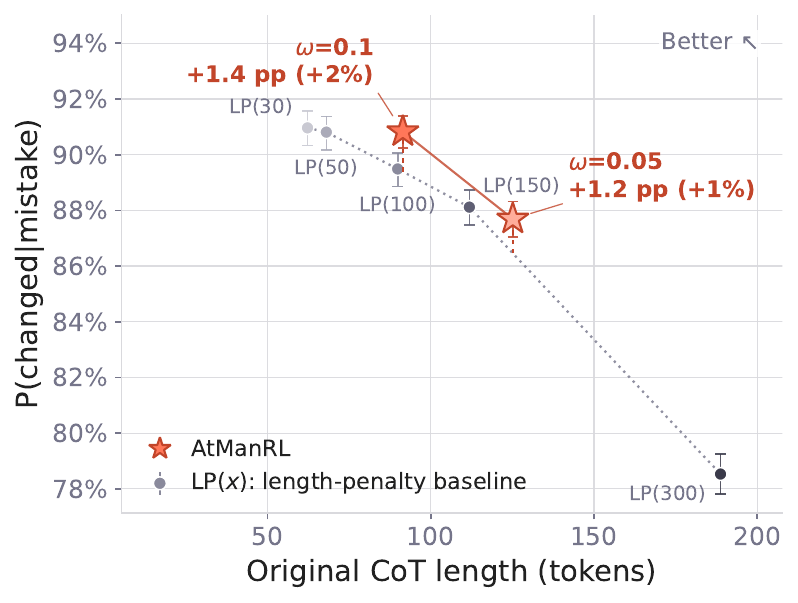}
    \caption{Mistake injection on GSM8K, pooled across models: change in CoT length.}
    \label{fig:mistake_dataset_gsm8k}
\end{figure}
\begin{figure}[H]
    \centering
    \includegraphics[width=0.99\linewidth, trim=0 0.2cm 0 0, clip]{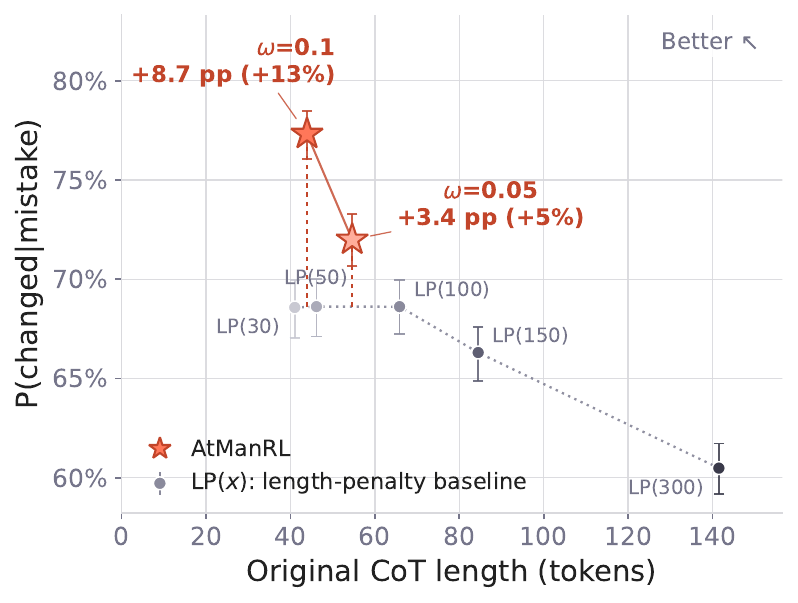}
    \caption{Mistake injection on QuestBench, pooled across models: change in CoT length.}
    \label{fig:mistake_dataset_questbench}
\end{figure}
\begin{figure}[H]
    \centering
    \includegraphics[width=0.99\linewidth, trim=0 0.2cm 0 0, clip]{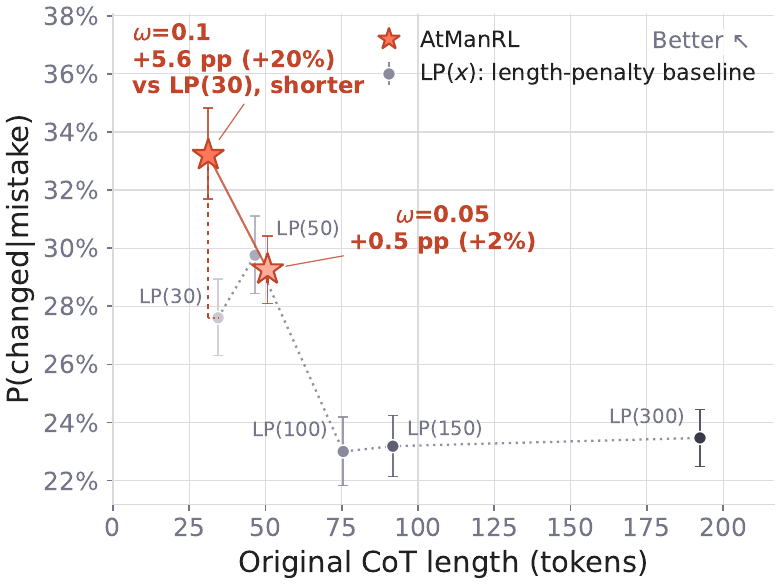}
    \caption{Mistake injection on MMLU, pooled across models: change in CoT length.}
    \label{fig:mistake_dataset_mmlu}
\end{figure}
\begin{figure}[H]
    \centering
    \includegraphics[width=0.99\linewidth, trim=0 0.2cm 0 0, clip]{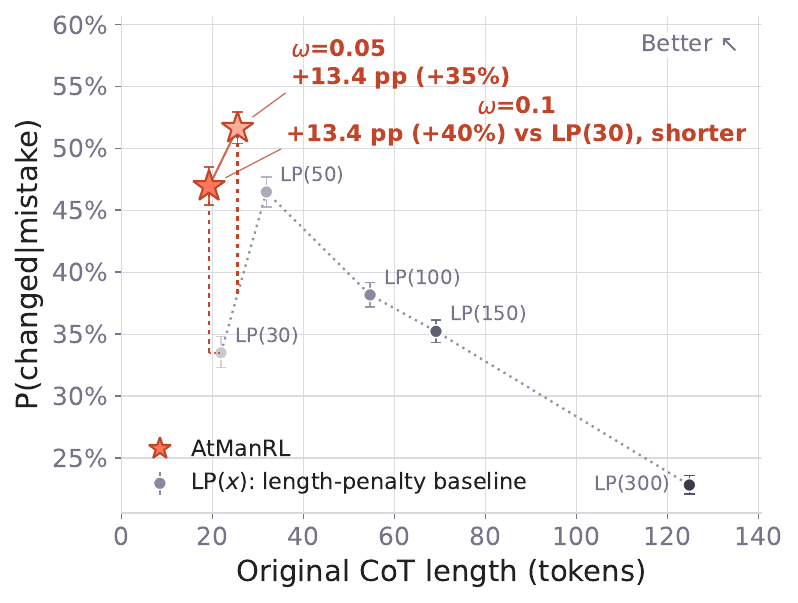}
    \caption{Mistake injection on ProofWriter, pooled across models: change in CoT length.}
    \label{fig:mistake_dataset_proofwriter}
\end{figure}
\subsection{Per-model}
\begin{figure}[H]
    \centering
    \includegraphics[width=0.99\linewidth, trim=0 0.2cm 0 0, clip]{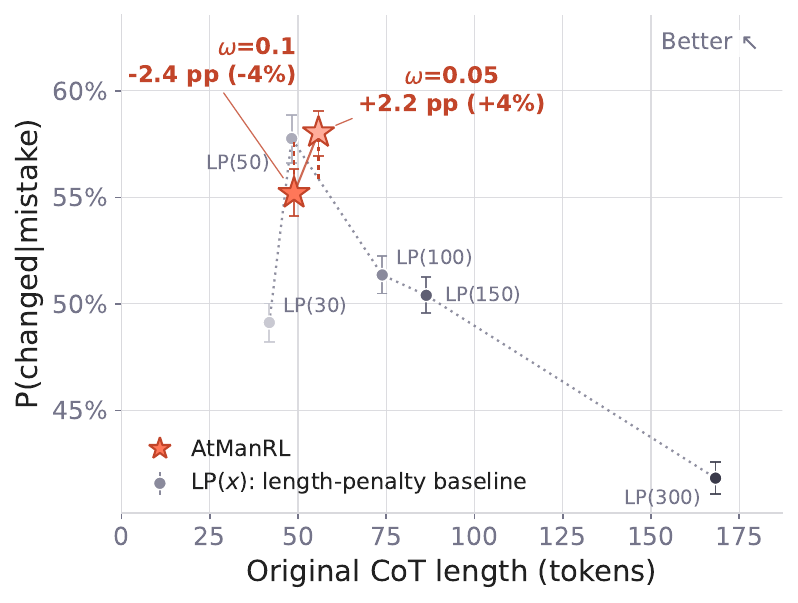}
    \caption{Mistake injection on Qwen2.5, pooled across datasets: change in CoT length.}
    \label{fig:mistake_model_qwen25}
\end{figure}
\begin{figure}[H]
    \centering
    \includegraphics[width=0.99\linewidth, trim=0 0.2cm 0 0, clip]{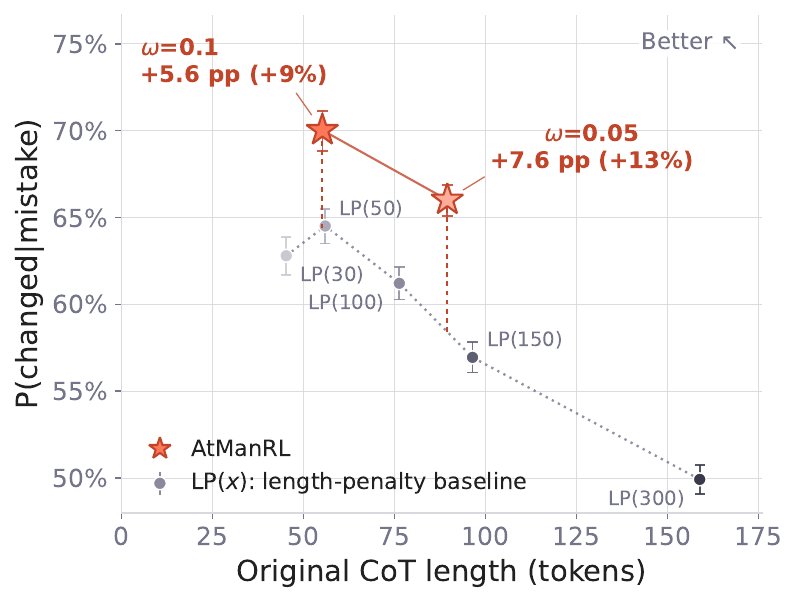}
    \caption{Mistake injection on Granite-3, pooled across datasets: change in CoT length.}
    \label{fig:mistake_model_granite3}
\end{figure}
\begin{figure}[H]
    \centering
    \includegraphics[width=0.99\linewidth, trim=0 0.2cm 0 0, clip]{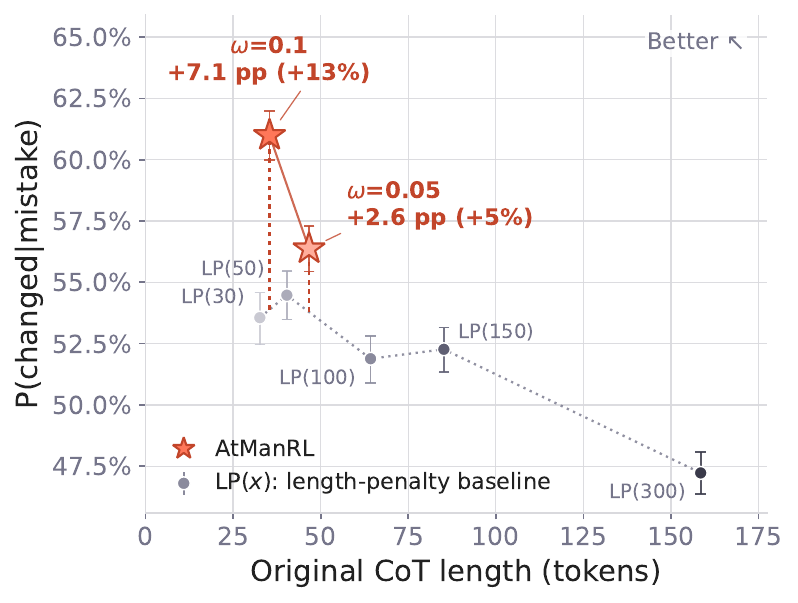}
    \caption{Mistake injection on Phi-4, pooled across datasets: change in CoT length.}
    \label{fig:mistake_model_phi4}
\end{figure}

\end{document}